\documentclass{article}
\usepackage{kr}

\usepackage{times}
\usepackage{soul}
\usepackage{url}
\usepackage[hidelinks]{hyperref}
\usepackage[utf8]{inputenc}
\usepackage[small]{caption}
\usepackage{graphicx}
\usepackage{amsmath}
\usepackage{amsthm}
\usepackage{booktabs}
\usepackage{algorithm}
\usepackage[noend]{algpseudocode}
\algrenewcommand\algorithmicdo{}
\algrenewcommand\algorithmicthen{}
\usepackage{latexsym}

\usepackage{amssymb}
\usepackage{bbold}
\usepackage{comment}
\usepackage{marvosym}
\usepackage{subcaption}
\usepackage{tikz}
\usetikzlibrary{arrows.meta,automata,calc,fit,matrix,patterns,positioning,shadows,shapes.arrows,shapes.geometric,svg.path}
\usepackage[marvosym]{tikzsymbols}
\usepackage[pro]{fontawesome5}
\usepackage{todonotes}
\usepackage{pifont}
\usepackage{cleveref}

\newtheorem{example}{Example}

\newtheorem{definition}{Definition}

\newcommand{\BibTeX}{B\kern-.05em{\sc i\kern-.025em b}\kern-.08em\TeX}

\newcommand{\inlinecite}[1]{\citeauthor{#1}~(\citeyear{#1})}

\newcommand{\set}[1]{\{#1\}}

\newcommand{\simplex}[1]{\Delta(#1)}
\newcommand{\arbitraryset}{\mathcal{X}}
\newcommand{\empseq}[1]{{#1}^\ast}
\newcommand{\nonempseq}[1]{{#1}^+}
\newcommand{\powerset}[1]{2^{#1}}

\newcommand{\event}[1]{{#1}=\top}
\newcommand{\notevent}[1]{{#1}=\bot}
\newcommand{\eventdetected}[1]{\Tilde{#1}=\top}
\newcommand{\noteventdetected}[1]{\Tilde{#1}=\bot}

\newcommand{\trace}{\lambda}
\newcommand{\noisytrace}{\tilde{\lambda}}
\newcommand{\states}{\mathcal{S}}
\newcommand{\state}{s}
\newcommand{\stateterm}{\state^T}
\newcommand{\stategoal}{\state^G}

\newcommand{\actions}{\mathcal{A}}
\newcommand{\action}{a}
\newcommand{\rewardfunction}{r}
\newcommand{\termfunction}{\tau}
\newcommand{\discountfactor}{\gamma}
\newcommand{\propositions}{\mathcal{P}}
\newcommand{\labelingfunction}{\mathcal{L}}
\newcommand{\noisylabelingfunction}{\widetilde{\mathcal{L}}}
\newcommand{\rmstates}{U}
\newcommand{\rmstate}{u}
\newcommand{\transitionprob}{p}
\newcommand{\rmstatebelief}{\tilde{u}}
\newcommand{\initialrmstate}{\rmstate_0}
\newcommand{\acceptingrmstate}{\rmstate_A}

\newcommand{\rejectingrmstate}{\rmstate_R}

\newcommand{\rmtransitionfunction}{\delta_\rmstate}
\newcommand{\observationlabel}{L}
\newcommand{\rmreward}{\delta_\rewardfunction}

\newcommand{\probability}[1]{P(#1)}

\newcommand{\history}{h}
\newcommand{\rewardmachine}{M}
\newcommand{\rmstatetransitionfun}{\delta_u}
\newcommand{\rmrewardtransitionfun}{\delta_r}
\newcommand{\policy}{\pi}
\newcommand{\step}{t}
\newcommand{\qfunc}{q}
\newcommand{\lr}{\alpha}
\newcommand{\rmtraversal}[2]{{#1}(#2)}  

\newcommand{\thesystem}{\mbox{\textsc{Prob-IRM}}\xspace}

\newcommand{\exampleset}{E}
\newcommand{\ilaspbackground}{B}
\newcommand{\modebias}{S_M}
\newcommand{\rmrelearningthreshold}{\beta}

\newcommand{\false}{\bot}
\newcommand{\true}{\top}

\newcommand{\ow}{\textsc{OfficeWorld}\xspace}
\newcommand{\owcoffee}{\textnormal{\Coffeecup}\xspace}
\newcommand{\owagent}{\Strichmaxerl[1.25]}
\newcommand{\owcella}{\ensuremath{A}\xspace}
\newcommand{\owcellb}{\ensuremath{B}\xspace}
\newcommand{\owcellc}{\ensuremath{C}\xspace}
\newcommand{\owcelld}{\ensuremath{D}\xspace}
\newcommand{\owdecor}{\ensuremath{\ast}\xspace}
\newcommand{\owoffice}{\ensuremath{o}\xspace}
\newcommand{\owmail}{\textnormal{\Letter}\xspace}
\newcommand{\owcoffeetask}{\textsc{Coffee}\xspace}
\newcommand{\owcoffeemailtask}{\textsc{CoffeeMail}\xspace}
\newcommand{\owvisit}{\textsc{VisitABCD}\xspace}

\definecolor{gold}{RGB}{255,215,0}
\definecolor{goldenrod}{RGB}{218,165,32}
\definecolor{maroon}{rgb}{0.5, 0.0, 0.0}

\def\stone(#1)[#2][#3][#4]{
    \begin{scope}[shift={(#1)},rotate=#2,scale=#3, transform shape]
        \draw[myshade,opacity=#4] (15:3pt) -- (45:4pt) -- (89:5pt)--(135:6pt) -- (150:3pt) -- (182:4pt)--(235:6pt) -- (280:3pt) -- (330:2pt) -- cycle;
    \end{scope}
}

\def\mine(#1)[#2]{
    \begin{scope}[shift={(#1)},transform shape]
        \stone(7pt,8pt)[55][1.1][#2]
        \stone(0pt,0pt)[45][0.7][#2]
        \stone(14pt,14pt)[60][0.7][#2]
        \stone(0pt,15pt)[110][0.5][#2]
        \stone(15pt,2pt)[-20][0.7][#2]
        \stone(7.5pt,0pt)[90][0.5][#2]
    \end{scope}
}

\tikzset{
	in place/.style={
		auto=false,
		fill=white,
		inner sep=2pt,
	},
	cascaded/.style = {%
		general shadow = {%
			shadow scale = 1,
			shadow xshift = -1ex,
			shadow yshift = 1ex,
			draw,
			fill = white},
		general shadow = {%
			shadow scale = 1,
			shadow xshift = -.5ex,
			shadow yshift = .5ex,
			draw,
			fill = white},
		fill = white, 
		draw,
		minimum width = 1.5cm,
		minimum height = 2cm
	},
	none/.style = {
		inner sep=0mm
	},
	every loop/.append style={-Latex}
}

\newcommand{\ilaspexample}{e}

\newcommand{\singleproposition}{l}

\newcommand{\bernoulli}[1]{Ber(#1)}

\title{Learning Robust Reward Machines from Noisy Labels\footnote{Accepted to the 21st International Conference on Principles of Knowledge Representation and Reasoning (KR2024)}
}

\author{%
Roko Parać$^1$\and
Lorenzo Nodari$^2$\and
Leo Ardon$^1$\and
Daniel Furelos-Blanco$^1$\and \\
Federico Cerutti$^{2,3}$\and
Alessandra Russo$^1$ \\
\affiliations
$^1$Imperial College London\\
$^2$University of Brescia\\
$^3$Cardiff University
\emails
roko.parac18@ic.ac.uk,
lorenzo.nodari@unibs.it,
leo.ardon19@ic.ac.uk,
d.furelos-blanco18@ic.ac.uk,
federico.cerutti@unibs.it,
a.russo@ic.ac.uk
}

\begin{document}

\maketitle


\begin{abstract}
This paper presents \thesystem, an approach that learns robust reward machines (RMs) for reinforcement learning (RL) agents from noisy execution traces. The key aspect of RM-driven RL is the exploitation of a finite-state machine that decomposes the agent's task into different subtasks. \thesystem uses a state-of-the-art inductive logic programming framework robust to noisy examples to learn RMs from noisy traces using the Bayesian posterior degree of beliefs, thus ensuring robustness against inconsistencies. Pivotal for the results is the interleaving between RM learning and policy learning: a new RM is learned whenever the RL agent generates a trace that is believed not to be accepted by the current RM. To speed up the training of the RL agent, \thesystem employs a probabilistic formulation of reward shaping that uses the posterior Bayesian beliefs derived from the traces. Our experimental analysis shows that \thesystem can learn (potentially imperfect) RMs from noisy traces and exploit them to train an RL agent to solve its tasks successfully. Despite the complexity of learning the RM from noisy traces, agents trained with \thesystem perform comparably to agents provided with handcrafted RMs.
\end{abstract}
\section{Introduction}
Reinforcement learning (RL; \citeauthor{SuttonB18}~\citeyear{SuttonB18}) is a machine learning paradigm where agents learn to solve a task by interacting with an environment to maximize their cumulative reward. Significant advancements have demonstrated its potential to perform tasks as well as, if not better than, humans.
A famous example is AlphaGo \cite{SilverHMGSDSAPL16}, the first AI system to beat a Go world champion.
In addition, RL plays a key role in the training of Large Language Models, such as ChatGPT \cite{DBLP:journals/corr/abs-2303-08774}.

Many open challenges still need to be addressed to make RL more widely applicable to real-world problems. Among these are the ability to generalise and transfer across tasks, operate robustly in the presence of partial observability and noise, and make the learned policies interpretable~\cite{DBLP:journals/ml/Dulac-ArnoldLML21}. 

Reward machines (RMs; \citeauthor{ToroIcarteKVM18}~\citeyear{ToroIcarteKVM18}) are a recent mechanism for addressing some of these challenges. RMs are finite-state machines representing non-Markovian reward functions in terms of high-level propositional events. The RM structures the agent's task into sequences of intermediate abstract states that act as an external memory for the agent. 
This makes the reward Markovian, thus enabling the application of standard RL algorithms in non-Markovian reward settings. The RM structure facilitates task decomposition, making policy learning more efficient when rewards are sparse. Recent work has extended the applicability of RMs by learning them instead of handcrafting them~\cite{ToroIcarteWKVCM19,XuGAMNTW20,FurelosBlancoLJBR21,HasanbeigJAMK21}, and hierarchically composing them, for  reusability~\cite{FurelosBlancoLJBR23}. However, RM learning approaches assume a \emph{perfect labelling function}, a construct that enables agents to accurately observe the high-level events occurring in the environment. This assumption is \emph{unrealistic}; for instance, robot sensors, which resemble the role of the labelling function, are seldom perfect, due to environmental conditions, limitations in technology, and inherent inaccuracies.

We propose $\thesystem$ (\textbf{Prob}abilisitic \textbf{I}nduction of \textbf{R}eward \textbf{M}achines), a method for learning and exploiting RMs from noisy propositions perceived by an RL agent through a \emph{noisy labelling function}.
$\thesystem$ uses ILASP~\cite{law2015ilasp}, a state-of-the-art inductive logic programming system capable of learning from noisy examples. 
The learned RM is exploited by an RL algorithm that leverages the RM structure using a novel probabilistic reward-shaping mechanism based on an RM state belief. The $\thesystem$ algorithm interleaves the RL and RM learning processes, enabling the agent to immediately exploit the newly learned (possibly sub-optimal) RMs. 
A new RM is learned during the interleaving process when the currently used one does not recognise noisy traces. 

%
%
We evaluate $\thesystem$ in several existing grid-world problems with sparse non-Markovian rewards. 
We show \thesystem learns RMs from noisy traces that are exploited by an RL agent. 
Our results demonstrate that under a wide array of noise configurations, $\thesystem$ performs similarly to approaches where RMs are handcrafted. 
%

The paper is organised as follows. \Cref{sec:background} introduces the background of our work. \Cref{sec:methodology} describes $\thesystem$, including the problem formalization and the learning and exploitation of RMs from noisy traces. \Cref{experiments} presents our experimental results, \Cref{sec:related} discusses related work, and \Cref{sec:conclusions} concludes the paper with suggestions for future directions. 

\section{Background}
\label{sec:background}
In this section, we introduce the basic notions and terminology. Given a finite set $\arbitraryset$, $\simplex{\arbitraryset}$ is the probability simplex over $\arbitraryset$, $\empseq{\arbitraryset}$ denotes (possibly empty) sequences of elements from $\arbitraryset$, $\nonempseq{\arbitraryset}$ denotes non-empty sequences of elements from $\arbitraryset$, and $\powerset{\arbitraryset}$ is the power set of $\arbitraryset$.

\subsection{Reinforcement Learning}
\label{Reinforcementlearning}
We formalize RL tasks as \emph{labelled Markov decision processes} (MDPs; \citeauthor{FuT14}~\citeyear{FuT14}; \citeauthor{FurelosBlancoLJBR23}~\citeyear{FurelosBlancoLJBR23}). A MDP is a tuple $\langle\states,\actions,\transitionprob,\rewardfunction,\termfunction,\discountfactor,\propositions,\labelingfunction \rangle$ where $\states$ is a set of states, $\actions$ is a set of actions, $\transitionprob:\states\times\actions\to\simplex{\states}$ is a probability transition function, $\rewardfunction:\nonempseq{(\states\times\actions)}\times\states\to\mathbb{R}$ is a reward function, $\termfunction:\empseq{(\states\times\actions)\times\states}\to\set{\false,\true}\times\set{\false,\true}$ is a termination function, $\discountfactor\in[0,1)$ is a discount factor, $\propositions$ is a finite set of \emph{propositions} representing high-level events, and $\labelingfunction:\states\times\actions\times\states\to\powerset{\propositions}$ is a \emph{(perfect) labelling function} mapping state-action-state triplets into sets of propositions. 
We refer to these sets as \emph{labels}. The transition function $\transitionprob$ is Markovian, whereas the reward function $\rewardfunction$ and the termination function $\termfunction$ are not (i.e.,~they are history-dependent). 

Given a state-action \emph{history} $\history_\step=\langle \state_0,\allowbreak\action_0,\allowbreak\ldots,\allowbreak\state_\step\rangle \in \empseq{(\states\times\actions)}\times\states$, a \emph{trace} $\trace_\step=\langle\labelingfunction(\emptyset,\emptyset,\state_0),\allowbreak\ldots,\allowbreak\labelingfunction(\state_{\step-1},\action_{\step-1},\state_\step)\rangle\in\nonempseq{(\powerset{\propositions})}$ assigns a label to all triplets in $\history_\step$. The goal is to find a \emph{policy} $\policy:\nonempseq{(\powerset{\propositions})}\times\states\to\simplex{\actions}$ that maps traces-states to a probability distribution over actions that maximizes the expected cumulative discounted reward (or \emph{return}) $R_\step=\mathbb{E}_\policy[\sum^n_{k=\step}\discountfactor^{k-\step}\rewardfunction(\history_\step)]$, where $n$ is the episode's last step. Traces must be faithful representations of history to find such a policy, i.e., reward and termination functions could depend on traces instead of history.

The agent-environment interaction is as follows. At time $\step$, the (label) trace is $\trace_\step\in\nonempseq{(\powerset{\propositions})}$ and the agent observes a tuple $\langle \state_\step, \stateterm_\step, \stategoal_\step\rangle$, where $\state_\step\in\states$ is the state, $\stateterm_\step\in\set{\false,\true}$ indicates whether the history is terminal, and $\stategoal_\step\in\set{\false,\true}$ indicates whether the history accomplishes the task's goal. Both $\stateterm_\step$ and $\stategoal_\step$ are determined by the termination function $\termfunction$. The agent also observes a label $\observationlabel_\step=\labelingfunction(\state_{\step-1},\action_{\step-1},\state_\step)$. If the history is non-terminal, the agent runs an action $\action_\step\in\actions$, and the environment transitions to state $\state_{\step+1} \sim \transitionprob(\cdot\mid\state_\step,\action_\step)$. The agent then observes a new tuple $\langle \state_{\step+1}, \stateterm_{\step+1}, \stategoal_{\step+1}\rangle$ and label $\observationlabel_{\step+1}$, extends the trace as $\trace_{\step+1}=\trace_\step \oplus \observationlabel_{\step+1}$, and receives reward $\rewardfunction_{\step+1}$. A trace $\trace_\step$ is a \emph{goal trace} if $\langle\stateterm_\step,\stategoal_\step\rangle=\langle\true,\true\rangle$, a \emph{dead-end trace} if $\langle\stateterm_\step,\stategoal_\step\rangle=\langle\true,\false\rangle$, and an \emph{incomplete trace} if $\stateterm_\step=\false$.

\begin{figure*}
    \centering
    \begin{subfigure}[b]{0.495\linewidth}
        \tikzset{digit/.style = { minimum height = 5mm, minimum width=5mm, anchor=center }}
        \newcommand{\setcell}[3]{\edef\x{#2 - 0.5}\edef\y{9.5 - #1}\node[digit,name={#1-#2}] at (\x, \y) {#3};}
        \centering
        \resizebox{0.6\linewidth}{!}{
        \begin{tikzpicture}[scale=0.5]
            \draw[white] (0, 0) grid (13, 10);
            
            \setcell{9}{2}{0} \setcell{9}{3}{1} \setcell{9}{4}{2} \setcell{9}{5}{3} \setcell{9}{6}{4} \setcell{9}{7}{5} \setcell{9}{8}{6} \setcell{9}{9}{7} \setcell{9}{10}{8} \setcell{9}{11}{9} \setcell{9}{12}{10} \setcell{9}{13}{11}
            
            \setcell{8}{1}{0} \setcell{7}{1}{1} \setcell{6}{1}{2} \setcell{5}{1}{3} \setcell{4}{1}{4} \setcell{3}{1}{5} \setcell{2}{1}{6} \setcell{1}{1}{7} \setcell{0}{1}{8}
            
            \draw[gray] (1, 1) grid (13, 10);
            \draw[very thick, scale=3] (1/3, 1/3) rectangle (13/3, 10/3);
            
            \draw[very thick, scale=1] (4, 1) rectangle (4, 2); \draw[very thick, scale=1] (4, 9) rectangle (4, 10);
            \draw[very thick, scale=1] (4, 3) rectangle (4, 8);
            \draw[very thick, scale=1] (7, 1) rectangle (7, 2); \draw[very thick, scale=1] (7, 3) rectangle (7, 8); \draw[very thick, scale=1] (7, 9) rectangle (7, 10);
            \draw[very thick, scale=1] (10, 1) rectangle (10, 2); \draw[very thick, scale=1] (10, 3) rectangle (10, 8); \draw[very thick, scale=1] (10, 9) rectangle (10, 10);
            \draw[very thick, scale=1] (1, 4) rectangle (2, 4); \draw[very thick, scale=1] (3, 4) rectangle (11, 4); \draw[very thick, scale=1] (12, 4) rectangle (13, 4);
            \draw[very thick, scale=1] (1, 7) rectangle (2, 7); \draw[very thick, scale=1] (3, 7) rectangle (5, 7); \draw[very thick, scale=1] (6, 7) rectangle (8, 7); \draw[very thick, scale=1] (9, 7) rectangle (11, 7); \draw[very thick, scale=1] (12, 7) rectangle (13, 7);
            \setcell{1}{3}{\owcelld} \setcell{1}{6}{\owdecor} \setcell{1}{9}{\owdecor} \setcell{1}{12}{\owcellc}
            \setcell{2}{5}{\owcoffee} \setcell{2}{6}{\owagent}
            \setcell{4}{3}{\owdecor} \setcell{4}{6}{\owoffice} \setcell{4}{9}{\owmail} \setcell{4}{12}{\owdecor}
            \setcell{6}{10}{\owcoffee}
            \setcell{7}{3}{\owcella} \setcell{7}{6}{\owdecor} \setcell{7}{9}{\owdecor} \setcell{7}{12}{\owcellb}
        \end{tikzpicture}
        }
    \end{subfigure}
    \hfill
    \begin{subfigure}[b]{0.495\linewidth}
        \centering
        \resizebox{0.8\linewidth}{!}{
	\begin{tikzpicture}[shorten >=1pt,node distance=2.06cm,on grid,auto,every initial by arrow/.style ={-Latex}]
            \node[state,initial,initial text=] (u_0)   {$\initialrmstate$};
            \node[state,accepting] (u_acc) [below =3cm of u_0]  {$\acceptingrmstate$};
            \node[state] (u_1) [left =3.5cm of u_acc]   {$\rmstate_1$};
            \node[state] (u_rej) [right =3.5cm of u_acc]  {$\rejectingrmstate$};
            
            \path[-Latex] (u_0) edge [loop above] node {$\langle \emph{o.w.}, 0 \rangle$} ();
            \path[-Latex] (u_1) edge [loop below] node {$\langle \emph{o.w.}, 0 \rangle$} ();
            \path[-Latex] (u_acc) edge [loop right] node {$\langle \emph{o.w.}, 0 \rangle$} ();
            \path[-Latex] (u_rej) edge [loop below] node {$\langle \emph{o.w.}, 0 \rangle$} ();
            
            \path[-Latex] (u_0) edge [bend right] node[in place] {$ \langle \owcoffee \land \neg \owoffice \land \neg \owdecor, 0 \rangle$} (u_1);
            \path[-Latex] (u_0) edge [bend left] node[in place] {$\langle \owdecor, 0 \rangle$} (u_rej);
            \path[-Latex] (u_1) edge[bend right] node[in place] {$\langle \owdecor, 0 \rangle$} (u_rej);
            \path[-Latex] (u_1) edge[bend left] node[in place] {$\langle \owoffice\land\neg\owdecor, 1 \rangle$} (u_acc);
            \path[-Latex] (u_0) edge node[in place] {$\langle \owcoffee \land \owoffice \land\neg\owdecor, 1 \rangle$} (u_acc);
         \end{tikzpicture}
         }
    \end{subfigure}
    \caption{An \ow instance (left) and a reward machine for the \owcoffeetask task (right), where \emph{o.w.} stands for \emph{otherwise}.}
    \label{fig:ow_rm}
\end{figure*}

\begin{example}
The \ow~\cite{ToroIcarteKVM18}, illustrated in Figure~\ref{fig:ow_rm} (left), is a $12\times 9$ grid labeled with some special locations. At each step, the agent $\owagent$ observes its current position in the grid and moves in one of the four cardinal directions; that is, $\states=\set{0,\ldots,11}\times\set{0,\ldots,8}$ and $\actions=\set{\textnormal{up},\textnormal{down},\textnormal{left},\textnormal{right}}$. The agent always moves in the intended direction and stays put if it moves towards a wall. The set of propositions $\propositions=\set{\owcoffee,\owmail,\owoffice,\owcella,\owcellb,\owcellc,\owcelld,\owdecor}$ is constituted of the special locations. The labelling function maps a $\langle s,a,s'\rangle\in\states\times\actions\times\states$ triplet to the set of propositions observed in $s'$, e.g.~$\labelingfunction((4,6),\textnormal{left},(3,6))=\set{\owcoffee}$. In this paper, we consider different tasks that consist of visiting a sequence of special locations while avoiding the decorations ($\owdecor$):
\begin{itemize}
    \item \owcoffeetask: go to the coffee machine (\owcoffee) then go to the office (\owoffice).
    \item \owcoffeemailtask: go to the coffee machine (\owcoffee) and the mail location (\owmail), in any order, then go to the office (\owoffice).
    \item \owvisit: go to locations \owcella, \owcellb, \owcellc and \owcelld in order.
\end{itemize}
A reward of 1 is obtained for completing the task; otherwise, the reward is 0.
Rewards are non-Markovian since the current state (i.e.,~position on the grid) alone cannot determine the reward. The history $\history=\langle\langle4,6\rangle,\allowbreak\textnormal{left},\allowbreak\langle3,6\rangle,\allowbreak\textnormal{right},\allowbreak\langle4,6\rangle,\allowbreak\textnormal{down},\allowbreak\langle4,5\rangle,\allowbreak\textnormal{down},\allowbreak\langle4,4\rangle\rangle$ yields a reward of 1 and is mapped to the goal trace $\lambda=\langle\set{},\allowbreak\set{\owcoffee},\allowbreak\set{},\allowbreak\set{},\allowbreak\set{\owoffice}\rangle$.
\end{example}

Learning policies over histories or traces is impractical since they can grow arbitrarily. In this paper, we employ \emph{reward machines} to compactly encode traces, enabling efficient policy learning.

\subsection{Reward Machines}
\label{sec:background_rms}
A \emph{reward machine} (RM; \citeauthor{ToroIcarteKVM18}~\citeyear{ToroIcarteKVM18,ToroIcarteKVM22}) is a finite-state machine representation of a reward function. Formally, an RM is a tuple $\rewardmachine=\langle\rmstates,\propositions,\rmstatetransitionfun,\rmrewardtransitionfun, \initialrmstate,\acceptingrmstate,\rejectingrmstate\rangle$, where $\rmstates$ is a set of states, $\propositions$ is a set of propositions constituting the RM's alphabet, $\rmstatetransitionfun:\rmstates\times\powerset{\propositions} \to \rmstates$ is a state-transition function, $\rmrewardtransitionfun:\rmstates\times\rmstates\to\mathbb{R}$ is a reward-transition function, $\initialrmstate\in\rmstates$ is the initial state, $\acceptingrmstate\in\rmstates$ is the accepting state, and $\rejectingrmstate\in\rmstates$ is the rejecting state.

\begin{example}
    Figure~\ref{fig:ow_rm} (right) illustrates the RM for the \ow's \owcoffeetask. The edges are labelled by propositional logic formulas over $\propositions=\set{\owcoffee,\owmail,\owoffice,\owcella,\owcellb,\owcellc,\owcelld,\owdecor}$ and rewards for transitioning between states. 
    To verify whether a formula is satisfied by a label $\observationlabel\in\powerset{\propositions}$, $\observationlabel$ is used as a truth assignment: propositions contained in the label are true, and false otherwise. For example, $\set{\owcoffee}\models \owcoffee\land\neg \owoffice\land\neg\owdecor$.
\end{example}

Reward machines are revealed to the agent during agent-environment interactions. Starting from the RM's initial state, the agent moves in the RM according to the state-transition function and obtains rewards through the reward-transition function. Given an RM $\rewardmachine$ and a trace $\trace=\langle\observationlabel_0,\ldots,\observationlabel_n\rangle$, a \emph{traversal} $\rmtraversal{\rewardmachine}{\trace}=\langle v_0, v_1, \ldots, v_{n+1} \rangle$ is a unique sequence of RM states such that (i)~$v_0=u_0$, and (ii)~$\rmtransitionfunction(v_i,\observationlabel_i)=v_{i+1}$ for $i=0,\ldots,n$. Traversals for goal and dead-end traces should terminate in the accepting and rejecting states, respectively; in contrast, traversals for incomplete traces should terminate somewhere different from the accepting and rejecting states.

\begin{example}
    Given the RM in Figure~\ref{fig:ow_rm} (right) and the goal trace $\lambda=\langle\set{},\allowbreak\set{\owcoffee},\allowbreak\set{},\allowbreak\set{},\allowbreak\set{\owoffice}\rangle$, the traversal is $\langle\initialrmstate,\rmstate_0,\rmstate_1,\rmstate_1,\rmstate_1,\acceptingrmstate\rangle$. As expected, the traversal ends with the accepting state.
    \label{ex_trace}
\end{example}

Reward machines constitute compact trace representations: each RM state encodes a different completion degree of the task. Consequently, rewards become Markovian when defined over $\states\times\rmstates$. In line with this observation, \inlinecite{ToroIcarteKVM22} propose an algorithm that learns an action-value function (or Q-function)---an estimation of the expected return following the execution of an action from a given state---over $\states\times\rmstates$.
Given a transition from state $\state$ to $\state'$ with an action $\action$ and its label $\observationlabel = \labelingfunction(\state, \action, \state')$, the Q-function $\qfunc:\states \times \rmstates \times \actions\to\mathbb{R}$ is updated as follows:
\begin{equation}
    \qfunc(\state, \rmstate,\action) \xleftarrow{\lr} \rmrewardtransitionfun(\rmstate,\rmstate')+\discountfactor\max_{\action'\in\actions}\qfunc(\state', \rmstate',\action'),
\end{equation}
where $\rmstate'=\rmtransitionfunction(\rmstate, \observationlabel)$, and $x\xleftarrow{\lr}y$ is shorthand for $x\leftarrow x+\lr(y-x)$. 
In the tabular case, where estimates are stored for each state-action pair, the algorithm converges to an optimal policy in the limit \cite[Theorem 4.1]{ToroIcarteKVM22}.

\subsection{Learning from Noisy Examples}
\label{sec:background_ilasp}
Previous work shows that RMs can be learned from traces during the RL agent's training~\cite{ToroIcarteWKVCM19,XuGAMNTW20,FurelosBlancoLJBR21,HasanbeigJAMK21}. We adopt a methodology similar to that by \citeauthor{FurelosBlancoLJBR21} (\citeyear{furelos2020induction,FurelosBlancoLJBR21}), who used a state-of-the-art inductive logic programming system to induce RMs represented as answer set programs (ASP; \citeauthor{gelfond2014knowledge}~\citeyear{gelfond2014knowledge}) that represent the RMs. In what follows, we describe the fundamentals of the learning system we use. We refer the reader to the work by \inlinecite{DBLP:phd/ethos/Law18} for details.

Learning from Answer Sets (LAS; \citeauthor{law2020ilasp}~\citeyear{law2020ilasp}) is a paradigm for learning answer set programs. A LAS task is a tuple $\langle B, S_M, E\rangle$ where $B$ is an ASP program
called background knowledge, $S_M$ is a set of rules known as the hypothesis space, and $E$ is a set of examples. The hypothesis space is specified through a set of mode declarations $M$, describing which predicates can appear in the head or body of a rule.

ILASP~\cite{law2015ilasp} is a state-of-the-art system for solving LAS tasks robust to \emph{noisy examples}~\cite{law2018inductive}. Noisy examples in ILASP are referred to as \emph{weighted context-dependent partial interpretations} (WCDPIs), each a tuple $\langle e_{id}, e_{pen}, \langle e^{inc},e^{exc}\rangle,e_{ctx}\rangle$, where $e_{id}$ is an identifier, $e_{pen}$ is a penalty denoting the level of certainty of an example which can be either a positive integer or infinite, $\langle e^{inc},e^{exc}\rangle$ is a pair of atom sets known as \emph{partial interpretation}, and $e_{ctx}$ is an ASP program known as \textit{context} that expresses example-specific information. 
A hypothesis $H \subseteq S_M$ covers (or \emph{accepts}) a WCDPI if there exists an answer set of $B \cup e_{ctx} \cup H$ that contains every atom in $e^{inc}$ and no atom in $e^{exc}$.\footnote{We later show a WCDPI construction for our method (see \Cref{wcdpi-example}).} 
Informally, the \emph{cost} of a hypothesis $H$ is the sum of the hypothesis length plus the penalties of the examples not accepted by $H$. Examples with an infinite penalty are not noisy and must be covered by the induced hypothesis. The \emph{goal} of a LAS task with noisy examples is to find a hypothesis that minimises the cost over a given hypothesis space for a given set of WCDPIs. This is formally defined as follows. 

\begin{definition}\label{def:lnas}
  A LAS task $T$ is a tuple of the form $\langle B, S_M, E\rangle$, where $B$ is an ASP program, $S_M$ is a
  hypothesis space and $E$ is a set of WCDPIs.
  Given a hypothesis $H \subseteq S_M$:
  \begin{enumerate}
    \item $uncov(H, T)$ is the set of
      all examples $e \in E$ such that $H$ does not accept $e$.
    \item
      the penalty of $H$, denoted as $pen(H, T)$, is the sum $\sum_{e \in
      uncov(H, T)} e_{pen}$.
    \item the length of $H$, denoted as $length(H)$, is the sum 
    $\sum_{r \in H} length(r)$, where $length(r)\in\mathbb{N}_{>0}$.
    \item
      the score of $H$, denoted as $\mathcal{S}(H, T)$, is the sum $length(H) + pen(H, T)$.
    \item $H$ is an inductive solution of $T$ iff $\mathcal{S}(H, T)$ is finite.
    \item $H$ is an \emph{optimal inductive solution} of $T$ iff $\mathcal{S}(H, T)$ is finite and
      $\nexists H' \subseteq S_M$ such that $\mathcal{S}(H, T) >
      \mathcal{S}(H', T)$.
  \end{enumerate}

\end{definition}

The length of a hypothesis $H$ is often defined as the number of literals that appear in $H$, thus considering $length(r) = |r|$. However, this aspect can be customised to the specific task in hand, particularly when the task does not require the shortest hypotheses to be learned.
\section{Methodology}
\label{sec:methodology}
In this section, we present $\thesystem$, an approach for interleaving the learning of RMs from noisy traces with the learning of policies to solve a given task. We consider episodic labelled MDP tasks where the reward is 1 if the agent observes a goal trace and $0$ otherwise. The noisy traces are caused by a \emph{noisy labelling function}, which plays the role of an imperfect set of sensors. To cater for this uncertainty, RM traversals must account for the agent's \emph{belief} of being at a certain RM state.   





\subsection{Noisy Traces}

We assume the agent has a binary sensor for each proposition $\singleproposition\in \propositions$ with a given sensitivity and specificity of detection. We denote with $\event{\singleproposition}$ (resp.~$\notevent{\singleproposition}$) the \emph{actual} occurrence (resp.~absence) of $\singleproposition$ in the environment. We denote with $\eventdetected{\singleproposition}$ (resp.~$\noteventdetected{\singleproposition}$) the detection (resp.~non-detection) of $\singleproposition$ with the agent's sensors. The \emph{sensitivity} of the sensor is the probability of the sensor detecting a proposition given that it occurred; formally, $\probability{\eventdetected{\singleproposition}\mid\event{\singleproposition}}$. The \emph{specificity} of the sensor is the probability of not detecting a proposition given that it did not occur; formally, $\probability{\noteventdetected{\singleproposition}\mid\notevent{\singleproposition}}$.

As the sensor's prediction may differ from the actual occurrence of a proposition, the agent's belief of the (non)occurrence of a proposition is a posterior probability conditional to its sensor's value. We define the \emph{posterior} probability of a proposition $l$ using Bayes' rule:
%
%
\begin{align*}
\probability{l=\top\mid\Tilde{l}=y} =
\frac{\probability{\Tilde{l}=y\mid l=\top}\probability{l=\top}}
{\sum_{x \in \set{\top, \bot}}
\probability{\Tilde{l}=y\mid l=x} \probability{l=x}}
\end{align*}
where:
\begin{itemize}
    \item $y \in \set{\top, \bot}$ is the sensor detection outcome.
    \item $\probability{\eventdetected{\singleproposition}\mid\notevent{\singleproposition}} = 1 -\probability{\noteventdetected{\singleproposition}\mid\notevent{\singleproposition}}$, and represents the probability of an unexpected detection. Similarly, $\probability{\noteventdetected{\singleproposition}\mid\event{\singleproposition}} = 1 - \probability{\eventdetected{\singleproposition}\mid\event{\singleproposition}}$. 
    
    \item $\probability{\event{\singleproposition}}$ is the prior probability of $l$ occurring, and $\probability{\notevent{\singleproposition}}=1-\probability{\event{\singleproposition}}$. We refer the reader to Section~\ref{experiments} for further details on defining the prior. 
\end{itemize}

The labelling function must report the sensors' degree of uncertainty on the detected propositions. We introduce the notion of \emph{noisy labelling function}, which defines (for a given transition) the probability of every proposition conditioned to detecting their respective sensor readings. First, we define the probability of a proposition at a given transition. Given a transition triplet $(s_t,a_t,s_{t+1})$, 
we denote with $\tilde{\propositions}_{t+1} \in  \powerset{\propositions}$ all sensors' detections (i.e. all detected propositions) at that transition.
We assume that the probability of a proposition $l$ at a given transition is conditional only to its related sensor detection at that transition ($\tilde{l}_{t+1}$). 
The value $\tilde{l}_{t+1}$ is $\top$ if the proposition $l \in \tilde{\propositions}_{t+1}$ (detected); otherwise $\tilde{l}_{t+1}$ is $\bot$. So we define $P(l=\top\mid s_t, a_t, s_{t+1}) = P(l=\top \mid \tilde{\propositions}_{t+1})=P(l=\top \mid\tilde{l}_{t+1})$.

The noisy labelling function defines the probability of each proposition at a given transition.
%

\begin{definition}[Noisy labelling function] 
\label{def:noisylabelfun}
Let $\states$ be the set of states, $\actions$ the set of actions, and $\propositions$ be the set of propositions. Given a transition $(s_t,a_t,s_{t+1})\in \states \times \actions \times \states$, the noisy labelling function $\noisylabelingfunction$ maps the transition to the set of all possible propositions with their respective probabilities at that transition. Formally,  
$\noisylabelingfunction(s_t,a_t,s_{t+1}) = \set{(\singleproposition, \probability{\event{\singleproposition} \mid s_t,a_t,s_{t+1}}) \mid \singleproposition \in \propositions})$.
\label{noisy-labeling-function}
\end{definition}


We can then define the probability of a set of propositions at a given transition in terms of the noisy labelling function. Propositions are assumed to be conditionally independent. We use $[\noisylabelingfunction(\state_t, \action_t, \state_{t+1})]_\singleproposition$  to denote the probability of a proposition $l$ at the transition ($\state_t, \action_t, \state_{t+1})$ given by our noisy labelling function. So, given a label $\observationlabel \in \powerset{\propositions}$, we have:
\begin{gather}
   \probability{\observationlabel|\state_t, \action_t, \state_{t+1}}
   = \prod_{\singleproposition \in \propositions} p_l,
\end{gather}
where
\begin{equation*}
     p_l = 
     \begin{cases}
     1 - [\noisylabelingfunction(\state_t, \action_t, \state_{t+1})]_\singleproposition &  
     \text{ if } \singleproposition \notin \observationlabel; \\
     [\noisylabelingfunction(\state_t, \action_t, \state_{t+1})]_\singleproposition &
     \text{ if } \singleproposition \in \observationlabel.
     \end{cases}
\end{equation*}


\begin{definition}[Noisy trace]
Given a state-action history $\history_t = \langle s_0, a_0, s_1, a_1, ..., s_t \rangle$, the noisy trace $\noisytrace_{t}$ is given by:
\begin{align*}
    \noisytrace_{t} &= \langle \noisylabelingfunction(\state_0, \action_0, \state_1), 
    \noisylabelingfunction(\state_1, \action_1, \state_2),
    ...,
    \noisylabelingfunction(\state_{\step - 1}, \action_{\step - 1}, \state_\step) \rangle.
\end{align*}
\end{definition}


Our RL task can be formalised as a \emph{noisy labelled MDP}; that is, a labelled MDP (see Section~\ref{Reinforcementlearning}) but with a noisy labelling function. Despite the labelling function now being noisy, we \emph{assume} the termination function $\termfunction$ remains deterministic; hence, the termination of a noisy trace is determined with certainty throughout the agent-environment interaction. 

\subsection{Learning RMs from Noisy Traces}

We learn candidate RMs using ILASP. Recall that a LAS task is a tuple $\langle \ilaspbackground, \modebias, \exampleset\rangle$, where $\ilaspbackground$ is the background knowledge, $\modebias$ is the hypothesis space, and $\exampleset$ is a set of WCDPI examples (see \Cref{sec:background_ilasp}). The background knowledge $\ilaspbackground$ and the hypothesis space $\modebias$ are similar to the ones proposed by \inlinecite{FurelosBlancoLJBR21}. The former is a set of ASP rules that describes the general behaviour of any RM (i.e., how an RM is traversed). The latter contains all possible rules that can constitute the state-transition function of the RM.

We now focus on representing the WCDPI examples in $E$ from a given set of noisy traces.


\subsubsection{Generating Examples from Noisy Traces.}
\label{example-explanation}

One of the key aspects of learning RMs from noisy traces is how to map these traces into WCDPIs. 
A direct approach involves aggregating the probabilities generated by the noisy labelling function, for a given noisy trace, into a single trace-level probability. 
This aggregated probability can then be used as the weight for the WCDPI generated from that trace.
Approaches along this line have been proposed in the literature for other domains \cite{DBLP:journals/ml/CunningtonLLR23}.
They define the aggregation function as a t-norm over the collection of probabilistic predictions. This value is used as the \emph{penalty} for the examples, while the predictions are converted into their most likely outcome and stored within the WCDPI \emph{context}. Such a solution proved too restrictive since such WCDPIs could only represent the most likely trace.
Instead, we adopt a \emph{sampling-based method}: the probabilities in a noisy trace are used to define proposition-specific Bernoulli distributions, which are then sampled to determine the propositions that would be part of the context of the associated WCDPI.

Let $\noisytrace$ be a noisy trace. We denote with $\noisytrace_{i, \singleproposition} = [\noisylabelingfunction(\state_{i-1}, \action_{i-1}, \state_i)]_\singleproposition$ the probability of proposition $\singleproposition$ occurring at step $i$ of $\noisytrace$. We generate a \emph{sample trace} $\lambda'$ from $\noisytrace$ by determining the occurrence of each proposition $l\in\propositions$ at each step $i$ using the corresponding Bernoulli distribution; that is, $\lambda'_{i, \singleproposition} \sim
\bernoulli{\noisytrace_{i, \singleproposition}}$. For each sampled trace $\lambda'$, an \emph{ASP trace} $\lambda^{ASP}=\{\mathtt{prop}(\singleproposition,i) \mid \lambda'_{i,\singleproposition} = 1 \}$ is built by mapping each occurring proposition into a $\mathtt{prop}(\singleproposition,i)$ fact indicating that proposition $\singleproposition$ occurs at time $i$. As proposed by \inlinecite{FurelosBlancoLJBR21}, we \emph{compress} sampled traces by removing consecutive occurrences of the same sampled proposition set.

The WCDPI example generated from $\noisytrace$ is given by $\langle e_{id}, e_{pen}, \langle e^{inc}, e^{exc}\rangle, e_{ctx}\rangle$, where $e_{id}$ is a unique example identifier, and the penalty is $e_{pen} = 1$.  The partial interpretation $\langle e^{inc}, e^{exc}\rangle$ is $\langle\{\mathtt{accept}\},\{\mathtt{reject}\} \rangle$ for goal traces, $\langle\{\mathtt{reject}\},\{\mathtt{accept}\} \rangle$ for dead-end traces, and $\langle\emptyset,\{\mathtt{accept, reject}\} \rangle$ for incomplete traces. The atom $\mathtt{accept}$ (resp.~$\mathtt{reject}$) indicates that the accepting (resp.~rejecting) state of the RM is reached; therefore, for instance, the partial interpretation for goal traces indicates that the accepting state must be reached, whereas the rejecting state must not. Finally, the context $e_{ctx}$ is given by the ASP representation $\lambda^{ASP}$ of the sample trace $\lambda'$.

\begin{example}[Generation of WCDPI from a noisy trace]
Let us consider the proposition set $\propositions = \set{\owcoffee, \owoffice}$ 
and the noisy goal trace $\noisytrace = 
\langle  
\set{\owcoffee\!:\! 0.01, \owoffice\!:\! 0.01},\allowbreak
\set{\owcoffee\!:\! 0.9, \owoffice\!:\! 0.01},\allowbreak
\set{\owcoffee\!:\! 0.9, \owoffice\!:\! 0.01},\allowbreak
\set{\owcoffee\!:\! 0.01, \owoffice\!:\! 0.01},\allowbreak
\set{\owcoffee\!:\! 0.01, \owoffice: 0.9}
\rangle
$.
A WCDPI generated from $\noisytrace$ will be 
$\langle id,1,
\langle\{\mathtt{accept}\}, \{\mathtt{reject}\}\rangle, e_{ctx}\rangle$, where $e_{ctx}$ is constructed as follows:
\begin{enumerate}
    \item Produce a sample trace, e.g.~$\lambda'= 
\langle 
\set{\owcoffee: 0, \owoffice: 0},
\set{\owcoffee: 1, \owoffice: 0},
\set{\owcoffee: 1, \owoffice: 0},
\set{\owcoffee: 0, \owoffice: 0},
\set{\owcoffee: 0, \owoffice: 1}
\rangle$.
    \item Compress the trace: $\lambda'= 
    \langle 
    \set{\owcoffee: 0, \owoffice: 0},
    \set{\owcoffee: 1, \owoffice: 0},
    \set{\owcoffee: 0, \owoffice: 0},
    \set{\owcoffee: 0, \owoffice: 1}
    \rangle$.
    \item Construct the ASP representation $\lambda^{ASP}=
    \{\mathtt{prop}(\owcoffee, 1), \allowbreak \mathtt{prop}(\owoffice, 3)$\} of the compressed sample trace $\lambda'$.
\end{enumerate}


\label{wcdpi-example}
\end{example}

Once all the WCDPI for all the noisy traces are sampled, we make the following adjustments to their penalties. First, we reweigh the penalties to balance the classes (goal, dead-end, incomplete). Second, since the sampling process may produce $x$ identical WCDPIs $\langle \ilaspexample_{id_i}, \ilaspexample_{pen}, \langle \ilaspexample^{inc}, \ilaspexample^{exc} \rangle, \ilaspexample_{ctx} \rangle$, we replace them (without loss of generality) with one WCDPI of the form
$\langle \ilaspexample_{id_{new}}, x \cdot \ilaspexample_{pen}, \langle \ilaspexample^{inc}, \ilaspexample^{exc} \rangle, \ilaspexample_{ctx} \rangle$. 

To ensure the RM is well-formed and to make the RM learning more efficient, we enforce the determinism and the symmetry-breaking constraints proposed by 
\inlinecite{FurelosBlancoLJBR21}.

\subsection{Exploitation of Reward Machines}
\label{subsec:rm_exploitation}
In this section, we describe how an RM is exploited to learn policies.  
\subsubsection{Reward Machine State Belief.}
Because of the noisy labelling function, the current RM state cannot be known: we can only determine the \emph{RM state belief} (\citeauthor{LiCVKTM22}~\citeyear{LiCVKTM22,li2024reward}).

\begin{definition}[RM state belief $\rmstatebelief_t$] 
\label{rm-state-belief}
The RM state belief $\rmstatebelief_t \in \Delta(\rmstates)$ 
is a categorical probability distribution expressing the RL agent's belief of being in an RM state $\rmstate$ at timestep $t$. Formally,
\begin{align*}
\rmstatebelief_0(\rmstate) &= 
\begin{cases}
   1 &\text{if } \rmstate = \initialrmstate; \\
   0 &\text{otherwise},
\end{cases}
\\
\rmstatebelief_{t+1}(\rmstate) &= \sum_{\substack{\rmstate_t \in \rmstates,\\ \observationlabel_t \in 2^{\propositions}}}P(\observationlabel_t |\state_t, \action_t, \state_{t+1}) \rmstatebelief_t(\rmstate_t) \mathbb{1}[\rmtransitionfunction(\rmstate_t, \observationlabel_t) = \rmstate],
\end{align*}
where $\rmtransitionfunction$ is the RM state-transition function, and $\mathbb{1}$ is the indicator function. 
\end{definition}



\subsubsection{Probabilistic Reward Shaping.}


Reward shaping aims to provide additional rewards to guide the agent towards completing a task. Previous works use the \emph{potential-based reward shaping} \cite{NgHR99}, which generates intermediate rewards from the difference in values of a \emph{potential function} $\Phi(s)$ over consecutive MDP states. Under this formulation, reward shaping does not shrink the set of optimal policies.

In the context of RM-based RL, \inlinecite{CamachoIKVM19} and \inlinecite{FurelosBlancoLJBR21} define the potential function $\Phi: \rmstates \rightarrow \mathbb{R}$ in terms of RM states. Formally,
\begin{align*}
&r_s(u, u') = \gamma\Phi(u') - \Phi(u),
\end{align*}
where $\gamma$ is the MDP's discount factor.

Given that the agent has access to the belief vector $\rmstatebelief_\step \in \Delta(\rmstates)$,
we propose the \emph{potential function on RM state beliefs} $\Tilde{\Phi} : \Delta(\rmstates) \rightarrow \mathbb{R}$, which is defined as the sum of every plausible RM state's potential weighted by its belief. Formally,
\[
\Tilde{\Phi}(\Tilde{u}_t) = \sum_{u \in U} \Tilde{u}_t(u)\Phi(u),
\]
where $\Phi : \rmstates \rightarrow \mathbb{R}$ is a potential function on RM states. The resulting reward-shaping function can thus be expressed as:
\begin{align*}
r_s(\Tilde{u}_t, \Tilde{u}_{t+1}) &= \gamma \Tilde{\Phi}(\Tilde{u}_{t+1}) - \Tilde{\Phi}(\Tilde{u}_t) \\
&= \sum_{u \in U} \gamma \Tilde{u}_{t+1}(u) \Phi(u) - \Tilde{u}_{t}(u)\Phi(u) \\
&= \sum_{u \in U} (\gamma \Tilde{u}_{t+1}(u) - \Tilde{u}_{t}(u)) \Phi(u).
\end{align*}
\inlinecite{EckSDK13} introduced a similar formulation in the context of state beliefs in partially observable MDPs.

Akin to \inlinecite{FurelosBlancoLJBR21}, we define the potential function on RM states following the intuition that the agent should be rewarded for getting closer to $\acceptingrmstate$. Formally,\[
\Phi(\rmstate) = |\rmstates| - d_{\min{}}(\rmstate, \acceptingrmstate),
\]
where $d_{\min{}}(\rmstate, \acceptingrmstate)$ is the minimum distance between $\rmstate$ and $\acceptingrmstate$. If $\acceptingrmstate$ is unreachable from $\rmstate$, then $d_{\min{}}(\rmstate, \acceptingrmstate)=\infty$.

\begin{example}
[Reward shaping in the \ow's \owcoffeetask task]

From the reward machine in Figure \ref{fig:ow_rm}, we obtain the following potential function $\Phi$:
\begin{equation}
  \begin{split}
    \Phi(\rmstate_0) &= 4 - 1 = 3,\\
    \Phi(\acceptingrmstate) &= 4 - 0 = 4,
  \end{split}
    \quad\quad
  \begin{split}
    \Phi(\rmstate_1) &= 4 - 1 = 3,\\
    \Phi(\rejectingrmstate) &= 4 -\infty \nonumber = -\infty.
  \end{split}
\end{equation}
Given $\gamma=0.9$ and the RM state beliefs $\rmstatebelief_\step = [1,0,0,0]^\top$ and 
$\rmstatebelief_{\step+1} = [0,0.5,0.5,0]^\top$, the shaped reward is:
\begin{align*}
r_s(\rmstatebelief_\step, \rmstatebelief_{\step+1})
 &= \sum_{u \in U} (\gamma \Tilde{u}_{t+1}(u) - \Tilde{u}_{t}(u)) \Phi(u) \nonumber \\
 &= (0.9 \cdot 0 - 1) \cdot 3
  + (0.9 \cdot 0.5 - 0) \cdot 3 \nonumber \\
 &  + (0.9 \cdot 0.5 - 0) \cdot 4
  + (0.9 \cdot 0 - 0) \cdot -\infty \nonumber \\
 &= 0.15. \nonumber
\end{align*}
\end{example}

\subsection{Interleaved Learning Algorithm}
We now describe $\thesystem$, our method for interleaving the learning of RMs from noisy traces with RL.  The pseudocode is shown in Algorithm \ref{alg:interleaving_pseudocode}. 

\begin{algorithm}[ht]
    \caption{$\thesystem$ algorithm}
    \label{alg:interleaving_pseudocode}
    \begin{algorithmic}[1]
        \State $\rewardmachine \leftarrow$ \textsc{InitRM}($\lbrace \initialrmstate, \acceptingrmstate, \rejectingrmstate \rbrace$)
        \State $\exampleset \leftarrow \lbrace \rbrace$ \Comment Set of noisy examples 
        \State $\mathtt{step\_cnt} \leftarrow 0$\Comment{Steps since last RM learning}
        \State $\mathtt{ce\_sum} \leftarrow 0$ \Comment{Cross entropy sum}

        \State \textsc{InitQFunction}($\rewardmachine$)
	\For{$ep \in \{1,\ldots, \mathtt{num\_episodes}\}$}
            \State $\state, \rmstatebelief_p \leftarrow \textsc{EnvInitialState}()$
            \State $\noisytrace\leftarrow \langle \rangle~; \mathtt{t} \leftarrow 0~; \mathtt{done} \leftarrow \bot$
            \While{$\mathtt{done} = \bot$}
		      \State $\action \leftarrow$ \textsc{GetAction}($\state$, $\rmstatebelief_p$)
		      \State $\state', \mathtt{done} \leftarrow$ \textsc{EnvStep}($\state, \action$)
            \State $\rmstatebelief_q \leftarrow$ \textsc{GetRMBelief}($\noisylabelingfunction(\state, \action, \state'),\rmstatebelief_p$)
		      \State \textsc{UpdateTrace}($\noisylabelingfunction(\state, \action, \state'), \noisytrace$) 
                \State \textsc{UpdateQFunction}($\state, \action, \state', \rmreward(\rmstatebelief_p, \rmstatebelief_q), \rmstatebelief_q$)
                \State $\state \leftarrow \state'~; \rmstatebelief_p \leftarrow \rmstatebelief_q~; \mathtt{t} \leftarrow \mathtt{t} + 1$
            \If{\textsc{IsTerminal}($\rewardmachine$, $\rmstatebelief_p$)} \textbf{or} $\mathtt{t} > \mathtt{max\_ep\_len}$
            \State $\mathtt{done} \leftarrow \top$ 
            \EndIf
            
            \EndWhile
            \State $\exampleset_{new} \leftarrow$ \textsc{GenerateExamples}($\noisytrace$)
            \State $\exampleset_{new\_inc} \leftarrow$ \textsc{GenerateIncExamples}($\noisytrace$)
            \State $\exampleset \leftarrow \exampleset\cup\exampleset_{new}\cup\exampleset_{new\_inc}$
            \State $\mathtt{step\_cnt} \leftarrow \mathtt{step\_cnt} + 1$
            \State $\mathtt{ce}\leftarrow\textsc{RecognizeBelief}(\rewardmachine, \noisytrace, \rmstatebelief_p)$            
            \State $\mathtt{ce\_sum} \leftarrow \mathtt{ce} + \mathtt{ce\_sum}$ 

            \If{\textsc{ShouldRelearn}($ep$, $\mathtt{ce\_sum}$, $\mathtt{step\_cnt}$)}
                \State $\rewardmachine \leftarrow\textsc{RelearnRewardMachine}(\exampleset)$
                \State $\mathtt{step\_cnt} \leftarrow 0$
                \State $\mathtt{ce\_sum} \leftarrow 0$
                \State \textsc{InitQFunction}($\rewardmachine$)
            \EndIf
        \EndFor

        \Function{RecognizeBelief}{$\rewardmachine$, $\noisytrace$, $\rmstatebelief_{\step+1}$}
            \State $\mathtt{expected\_belief} \leftarrow \textsc{TraceOutcome}(\noisytrace)$
            \State \Return $\textsc{CrossEntropy}(\rmstatebelief_{\step+1}, \mathtt{expected\_belief})$
        \EndFunction

        \Function{ShouldRelearn}{$ep$, $\mathtt{ce\_sum}$, $\mathtt{step\_cnt}$}
            \If{$\mathtt{step\_cnt} < \mathtt{warmup\_steps}$}
               \State \Return $\bot$
            \EndIf
            \State \Return $\mathtt{ce\_sum} / \mathtt{step\_cnt} < \rmrelearningthreshold$
        \EndFunction
    \end{algorithmic}
\end{algorithm}


Lines 1--5 initialise the candidate RM $\rewardmachine$, the set of noisy ILASP examples $\exampleset$, variables tracking RM relearning, and the Q-function. The algorithm is then executed for a fixed number of episodes. For each episode step, the agent executes an action in the environment (line 11), gets the new RM state belief $\rmstatebelief$ (line 12), and updates the noisy trace (line 13) and the Q-function (line 14).
The episode terminates if (i) the environment signals that the task has been completed (line 11); (ii) the most likely state of the RM is the accepting/rejecting state; or (iii) after a fixed number of steps (lines 16--17). 
After the episode terminates, the ILASP examples are updated (line 20).
This process differs from the original algorithm~\cite[see \Cref{example-explanation}]{FurelosBlancoLJBR21}. We generate incomplete examples (line~19) from the newly seen traces similarly to \inlinecite{ArdonFR23}.

The belief that the RM is correct is updated using the observed trace (lines 22--23). If the RM should be relearned (line 24), then ILASP is called (line 25), variables tracking the relearning condition are reset (lines 26--27), and the Q-function is reinitialised (line 28).

The function \textsc{RecognizeBelief} (lines 29--31) computes how well the current trace conforms to the candidate RM.
We assume that the ground-truth outcome of an episode (goal, dead-end, incomplete) is known with certainty. 
This outcome is a one-hot vector obtained through the \textsc{TraceOutcome} function (line 30).
On the other hand, given that we know the belief of being in an accepting 
($[\rmstatebelief_{t+1}]_{acc}$) and rejecting state
($[\rmstatebelief_{t+1}]_{rej}$), we can compute the predicted outcome: 
($[\rmstatebelief_{t+1}]_{acc}$, 
$[\rmstatebelief_{t+1}]_{rej}$,
1 
- $[\rmstatebelief_{t+1}]_{acc}$ 
- $[\rmstatebelief_{t+1}]_{rej}$). The \textsc{RecognizeBelief} returns the categorical cross-entropy between the predicted and ground truth outcome.

The function \textsc{ShouldRelearn} (lines 32--35) determines if the RM should be relearned.
We use the average cross-entropy between the expected state belief for the observed trace and the current RM state as decision criteria to trigger the relearning of the RM: if the cross-entropy falls under the hyperparameter $\rmrelearningthreshold$, a new RM is learned.  Intuitively, this metric can be seen as a way to evaluate how well the RM captures the traces that have been seen.
Also, the RM should not be relearned too often to prevent relearning with similar examples and allow the new RM to influence the generation of new examples.
Hence, the RM should not be relearned if a number of warm-up steps have not passed since the last relearning (lines 33--34).

\section{Experimental Results}
\label{experiments}
We evaluate $\thesystem$ using the \ow (see \Cref{Reinforcementlearning}). We aim to answer three research questions:
\begin{description}
    \item[RQ1:] Does $\thesystem$ successfully allow agents to be trained to complete their tasks?
    \item[RQ2:] In terms of agent performance, how do the RMs learned with $\thesystem$ compare with hand-crafted ones?
    \item[RQ3:] How sensitive is $\thesystem$ to different noisy settings? 
\end{description}
After a brief overview of our experimental setup, we start by comparing the performance of $\thesystem$ agents with a baseline composed of agents provided with handcrafted RMs that perfectly expose the structure of each task. Then, we focus our analysis on the impact of significantly higher noise to highlight the robustness of $\thesystem$ in such scenarios. Finally, we present the results of two ablation studies that evaluate (i)~the effectiveness of our reward-shaping scheme, and (ii)~the performance difference between our belief-based approach and thresholding for handling noisy labels.

The source code is available at \url{https://github.com/rparac/Prob-IRM}.

\subsection{Experimental Setup}
\subsubsection{Environment Configurations.}
We focus our analysis on the \ow domain presented in \Cref{Reinforcementlearning}. On the one hand, the \owcoffeetask, \owcoffeemailtask, and \owvisit tasks enable assessing $\thesystem$'s ability to learn good-quality RMs and policies in progressively harder scenarios. On the other hand, thanks to its adoption in other existing RM-based work~\cite{ToroIcarteKVM18,FurelosBlancoLJBR21,DBLP:conf/aips/DohmenTAB0V22}, this choice allows our results to be easily compared with similar research.

The complexity of solving any of the three tasks strongly depends on the  \ow layout. To account for this, we conducted each experiment over 3 sets of 10 random maps for \owcoffeetask and \owcoffeemailtask, and 3 sets of 50 random maps for \owvisit.

\subsubsection{Sensor Configurations.}
We experiment with multiple sensor configurations, each with a specific choice of:
\begin{itemize}
    \item \emph{noise targets}: the set of sensors subject to noise. Either only the one associated with the first event needed to solve the task (\emph{noise-first}),\footnote{In \owcoffeemailtask, there is not a single event that satisfies this criterion; thus, we apply the noise on both the \owcoffee and \owmail sensors.} or all of them (\emph{noise-all}); 
    \item \emph{noise level}: the sensor detection specificity and sensitivity. To reduce the number of possible configurations, we only consider scenarios where both parameters are set to the same value, and we will thus refer to them jointly as \emph{sensor confidence}.
\end{itemize}

While sensor confidence is the parameter we have direct control over in our experiments, its value is not very informative to a human reader, as its impact on the accuracy of an agent's sensor strongly depends on the prior probabilities of each label being detected. Therefore, we pick the values for this parameter in such a way as to determine specific values for the posterior probability of detecting any noisy label correctly. In particular, we experiment with three posterior values, each representing increasing noise levels: 0.9, 0.8 and 0.5. We also consider the absence of noise (the posterior equal to 1) as a baseline.

\subsubsection{Policy Learning.}
The agent policies are learned via the RL algorithm for RMs outlined in \Cref{sec:background_rms}. The Q-functions are stored as tables. To index a reasonably sized Q-function using an RM belief vector, we resort to \emph{binning}: beliefs are truncated to a fixed number of decimals, effectively resulting in close values being considered identical.

In terms of exploration, we rely on an $\epsilon$-greedy strategy. For all the experiments, we start from $\epsilon = 1$ and decay its value to $\epsilon = 0.1$ over 2000 agent steps. We then keep the parameter fixed for the rest of the training process.

We employ our proposed formulation of \emph{probabilistic reward shaping} in all experiments (both with fixed and learnt RMs) using  $\gamma = 0.99$.

\subsubsection{Performance Metrics.}
We record the \emph{undiscounted return} collected by each agent at the end of every episode. Then, we aggregate their values over the set of all agents trained in the context of a single experiment to compute their associated mean and standard deviation, and the various metrics needed to answer our research questions. The average results are represented through learning curves, each smoothed using a moving average with a window size of 100. The shaded areas reflect the standard deviation of the values computed using an identical sliding window.


\begin{figure*}[t]
    \centering
    \begin{subfigure}[b]{0.33\linewidth}
        \centering
        \caption{\owcoffeetask}
        \includegraphics[width=\linewidth]{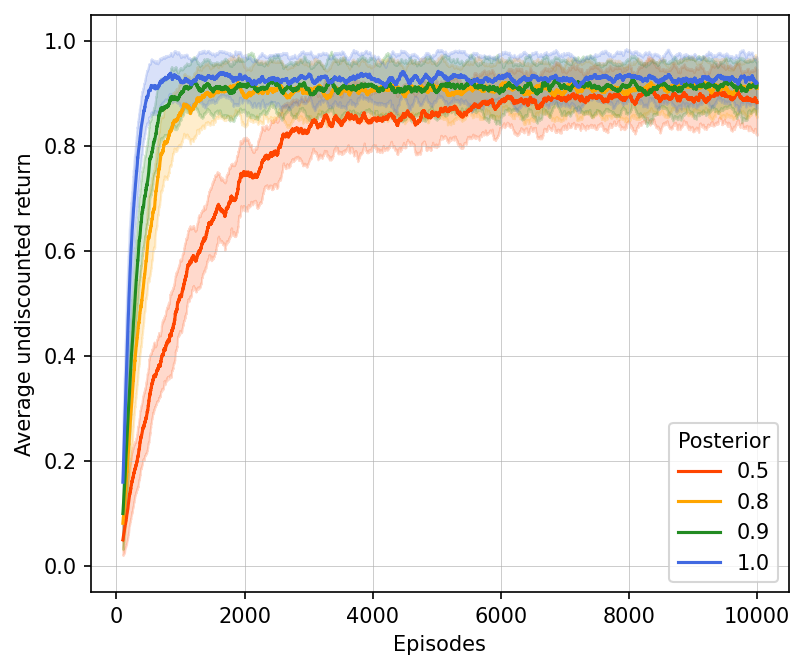}
        \label{fig:lc-coffee-fixedrm}
    \end{subfigure}
    \hfill
    \begin{subfigure}[b]{0.33\linewidth}
        \centering
        \caption{\owcoffeemailtask}
        \includegraphics[width=\linewidth]{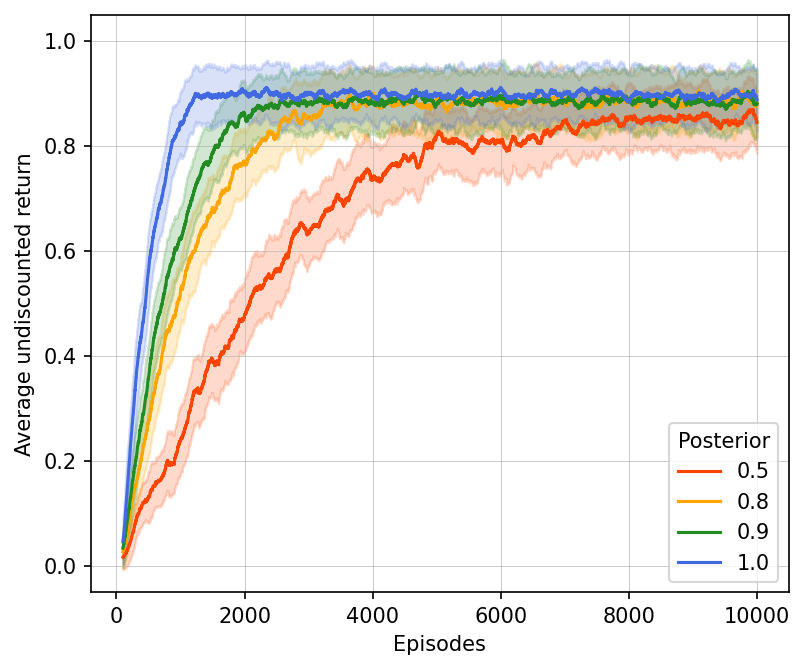}
        \label{fig:lc-mail-fixedrm}
    \end{subfigure}
    \hfill
    \begin{subfigure}[b]{0.33\linewidth}
        \centering
        \caption{\owvisit}
        \includegraphics[width=\linewidth]{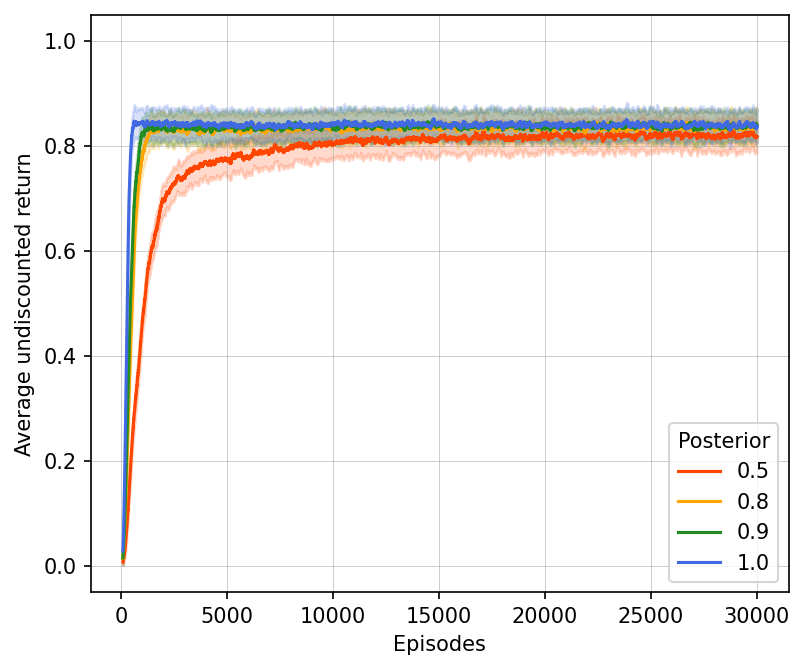}
        \label{fig:lc-abcd-fixedrm}
    \end{subfigure}
    \\
    \begin{subfigure}[b]{0.33\linewidth}
        \centering
        \includegraphics[width=\linewidth]{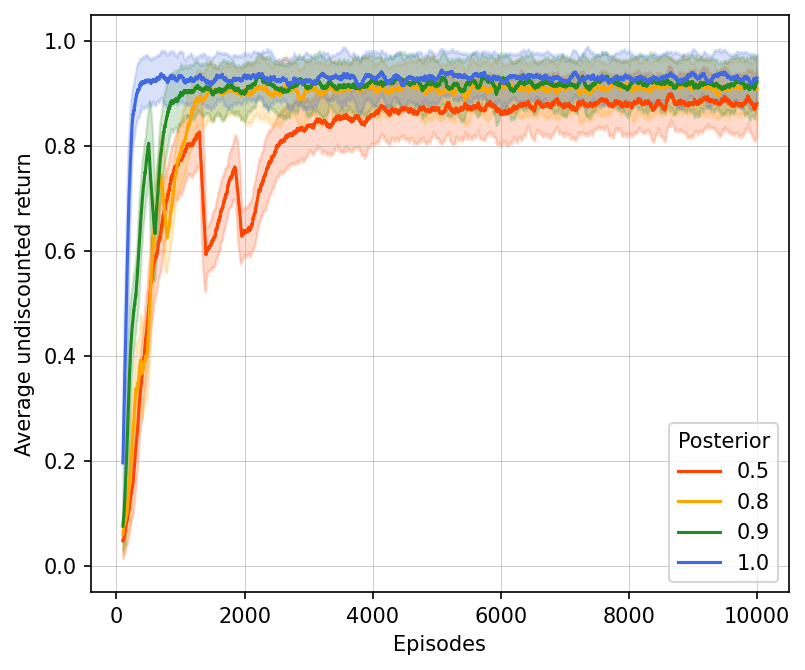}
        \label{fig:lc-coffee-probirm}
    \end{subfigure}
    \hfill
    \begin{subfigure}[b]{0.33\linewidth}
        \centering
        \includegraphics[width=\linewidth]{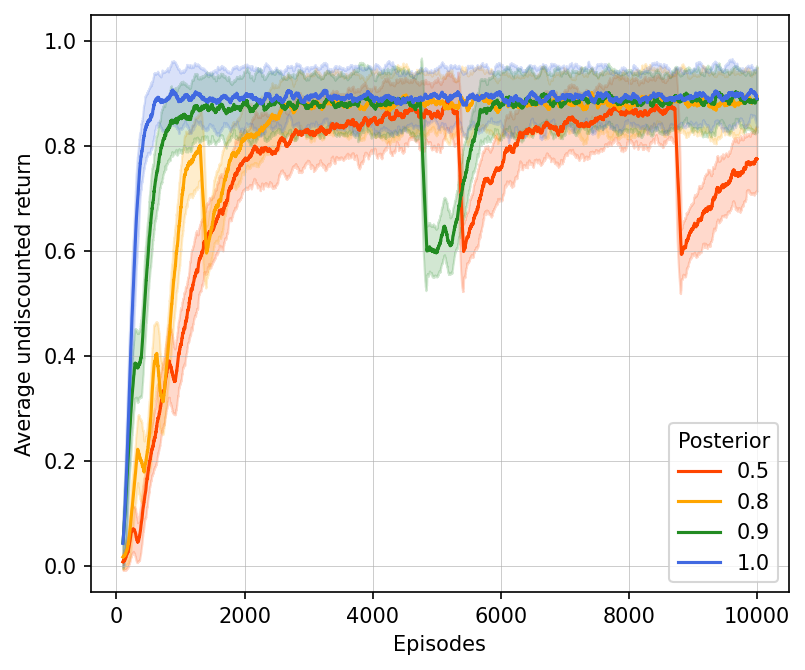}
        \label{fig:lc-mail-probirm}
    \end{subfigure}
    \hfill
    \begin{subfigure}[b]{0.33\linewidth}
        \centering
        \includegraphics[width=\linewidth]{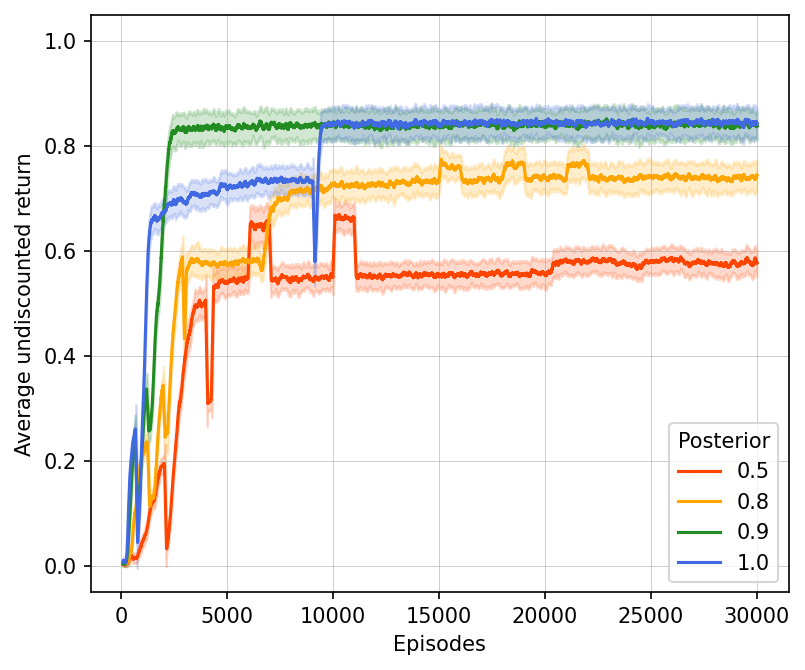}
        \label{fig:lc-abcd-probirm}
    \end{subfigure}
    \caption{Learning curves for the three \ow tasks, where the \emph{upper row} corresponds to the baseline agents provided with handcrafted RMs, and the \emph{lower row} corresponds to the agents trained with \thesystem.}
    \label{fig:baseline_results}
\end{figure*}

\subsection{Baseline Results}
We compare the performance of $\thesystem$ against a baseline composed of agents provided with handcrafted RMs. We focus on the \emph{noise-first} scenario, as it represents an easier learning setting for agents. 

The learning curves of both approaches are shown in \Cref{fig:baseline_results}. The baseline agents (upper row) consistently converge to a high level of proficiency. As expected, the more noise, the more episodes are required to converge.

We make two observations in the case of $\thesystem$ (lower row). First, most learning curves display one or more sudden changes: when a new RM is learnt, the agent's policy is discarded and a new one starts being trained, thus leading to a temporary decrease in performance. The frequency of RM relearning increases with the level of noise. Second, we observe that $\thesystem$ often reaches the baseline performance in a comparable number of episodes, thus providing a positive answer to research questions \textbf{RQ1} and \textbf{RQ2}. We highlight three exceptions: the first is found in the noisiest configuration for \owcoffeemailtask, where many agents triggered RM relearning near the end of their training, thus ending with subpar performance. The second and third exceptions are represented by the two most noisy configurations for \owvisit, which is the hardest task.

\subsection{Multiple Noise Sources}
\label{sec:mns}
We here assess $\thesystem$'s ability to deal with multiple independent sources of noise; hence, we consider the \emph{noise-all} setting for the \owcoffeetask task to answer \textbf{RQ3}.
In this setting, the number of noise sources is effectively tripled as compared to our baseline experiments: in addition to the noisy \owcoffee sensor, the \owdecor and \owoffice sensors are also noisy. 

We initially considered the set of posterior values used in the previous experiments; however, our experiments do not terminate within a 24-hour timeout period using a posterior of 0.5. This is unsurprising, as the learning task associated with this configuration is extremely complex. After experimenting with different posteriors ranging from 0.5 to 0.8, we find 0.75 to be a soft limit for the reliable applicability of $\thesystem$. Although more noise can be handled correctly, the training time increases rapidly. 

\Cref{fig:robustness_results} presents the results. We observe that $\thesystem$ agents consistently achieve a high performance despite the notable increase in noise. When comparing these results with the \emph{noise-first} scenario considered in the previous section, we notice a slight decrease in the average undiscounted return and a larger number of episodes required to converge. The magnitude of these effects appears to be directly proportional to the amount of noise on each sensor, as corroborated by the learning curve for a posterior of 0.75.

\subsection{Ablation Studies}
\subsubsection{Reward Shaping.}
\begin{figure}
    \centering
    \includegraphics[width=0.8\linewidth]{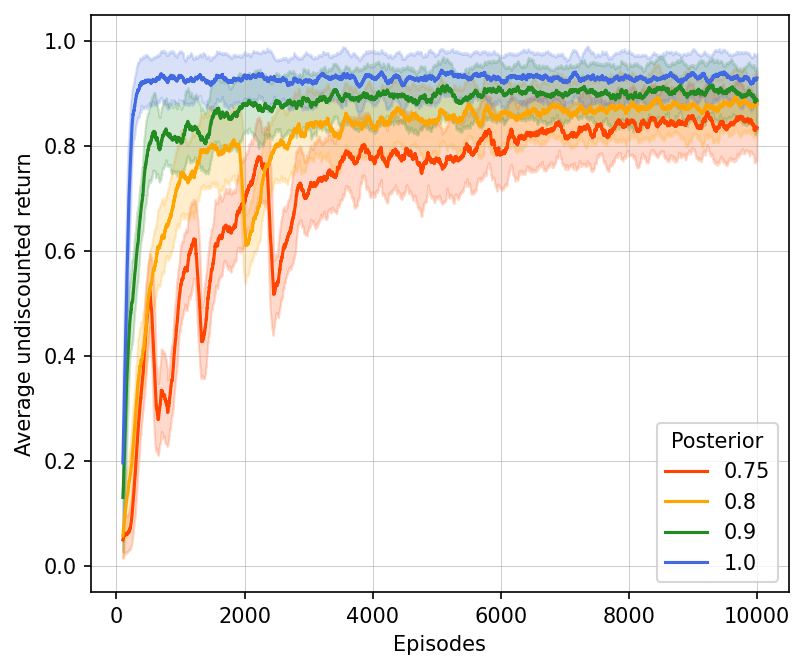}
    \caption{Learning curves for $\thesystem$ agents trained to solve the \owcoffeetask task in the \emph{noise-all} scenario.}
    \label{fig:robustness_results}
\end{figure}

\begin{figure}
    \centering
    \includegraphics[width=0.8\linewidth]{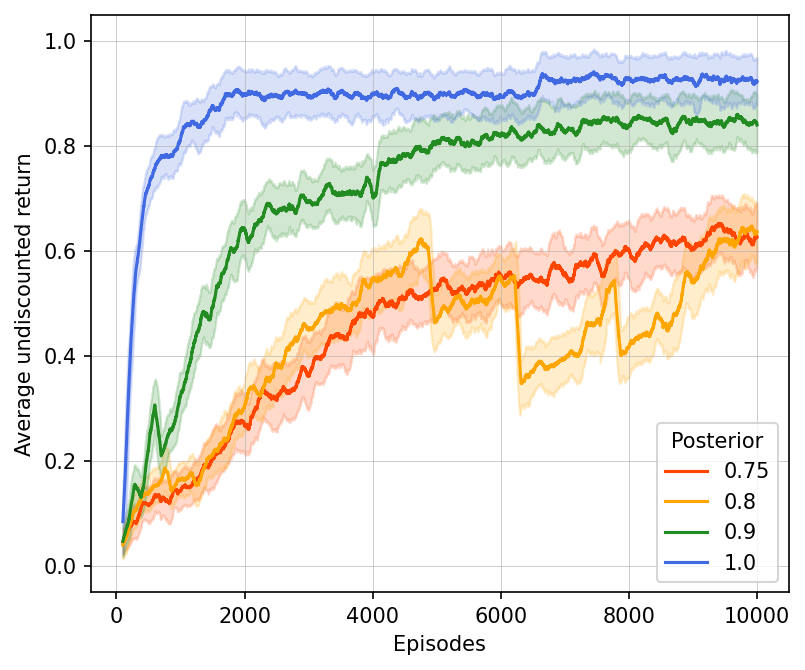}
    \caption{Learning curves for $\thesystem$ agents trained to solve the \owcoffeetask task in the \emph{noise-all} scenario \emph{without} reward shaping.}
    \label{fig:nors-results}
\end{figure}

To confirm the effectiveness of our reward shaping method, we replicate the setup in \Cref{sec:mns} but train the agents \emph{without} reward shaping. \Cref{fig:nors-results} presents the resulting learning curves. When comparing them with those in \Cref{fig:robustness_results}, the positive impact of reward shaping is immediately apparent: for each level of noise, its use leads to a substantial decrease in the number of training episodes required to reach high performance, in addition to an actual increase in the agents' proficiency at the end of training. Moreover, its effectiveness grows higher as the amount of noise in the environment increases, thus proving it to be a very useful tool for dealing with such scenarios. 

The reason behind its effectiveness lies in how reward shaping influences both policy training and RM learning. On the one hand, the intermediate rewards it provides help the agents improve their Q-value estimates faster, especially when rewards are sparse, thus leading to better policies sooner. On the other hand, they allow for incorrect RMs to be identified and discarded earlier in the learning process. When a candidate RM does not correctly reflect the task's structure, reward shaping nevertheless encourages the agents to act consequently. This, in turn, induces the agents to follow trajectories that are more likely to provide inconsistent examples for the RM learner.

\subsubsection{Post-Hoc Analysis: Belief Updating \emph{vs} Thresholding.}
In the last set of experiments, we aim to provide an empirical justification for our choice of a belief-updating paradigm, presented in \Cref{subsec:rm_exploitation}. A valid alternative is represented by \emph{thresholding}, which treats every proposition whose associated belief exceeds a fixed value as true. The main advantage of this strategy is not needing to maintain a belief over the RM states; instead, the RM state is updated by following the transitions triggered by the propositions deemed to be true after thresholding. Unfortunately, this upside is overshadowed by the fact that, under our sensor model, thresholding either works perfectly or does not work at all.

\begin{figure}
    \centering
    \includegraphics[width=0.8\linewidth]{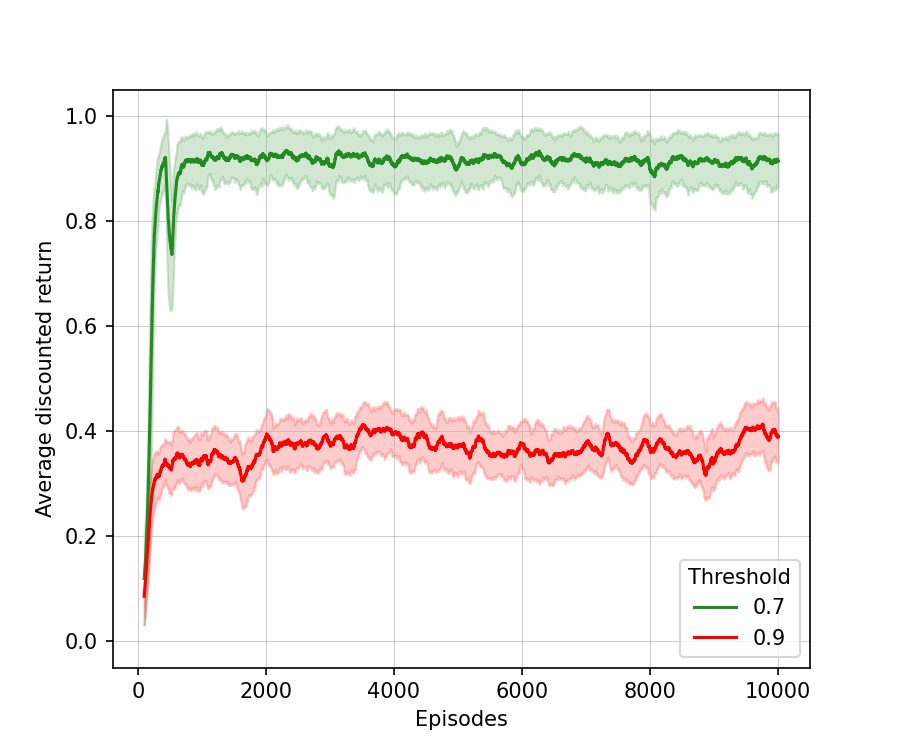}
    \caption{Learning curves for $\thesystem$ agents employing thresholding to solve the \owcoffeetask task under a \emph{noise-all} setting with a posterior of 0.8.}
    \label{fig:thresholding_comparison}
\end{figure}

To support this claim, we train two sets of $\thesystem$ agents to solve the \owcoffeetask task under the \emph{noisy-all} setting with a posterior of 0.8. The first set operates with a threshold of 0.7 (lower than the noise posterior), whereas the second uses a threshold of 0.9. Figure \ref{fig:thresholding_comparison} presents the resulting learning curves. We observe that thresholding achieves good results when the threshold is lower than the posterior since it recognizes the truth value of every proposition correctly; however, when the threshold is too high, the agents \emph{incorrectly} identify \emph{every} proposition, thus making learning impossible. This highlights the issue of thresholding under our sensor model: to be successful, thresholding requires either precise knowledge of the noise posterior an agent is subject to, or a careful tuning of the threshold parameter. Both requirements can be restrictive in many practical use cases. 

Unlike thresholding, the belief-updating approach we use is applicable regardless of the availability of any information, does not introduce an additional hyperparameter that might require extensive tuning, and performs very well in the same experimental setting we discuss in this section, as shown in \Cref{fig:robustness_results}.

\section{Related Work}
\label{sec:related}
Since the introduction of reward machines by \inlinecite{ToroIcarteKVM18}, there have been several approaches for learning them from traces in non-noisy settings. These approaches include discrete optimization~(\citeauthor{ToroIcarteWKVCM19}~\citeyear{ToroIcarteWKVCM19,ToroIcarteWKVCM23}), inductive logic programming~(ILP; \citeauthor{FurelosBlancoLJBR21}~\citeyear{FurelosBlancoLJBR21}; \citeauthor{ArdonFR23}~\citeyear{ArdonFR23}; \citeauthor{FurelosBlancoLJBR23}~\citeyear{FurelosBlancoLJBR23}), program synthesis~\cite{HasanbeigJAMK21}, or SAT solving~\cite{XuGAMNTW20,CorazzaGN22}. 
Our method employs ILASP \cite{law2015ilasp}, a system also used in the aforementioned ILP works, to learn the RMs. 
Given its inherent robustness to noisy examples, ILASP-based approaches were the best positioned to be extended.

Existing RM learning methods focus on either accurately predicting the next label from the previous one~\cite{ToroIcarteWKVCM19,HasanbeigJAMK21,ToroIcarteWKVCM23}, or learning a minimal RM (i.e.,~with the fewest number of states) that makes the reward signal Markovian~\cite{XuGAMNTW20,FurelosBlancoLJBR21,HasanbeigJAMK21,CorazzaGN22,ArdonFR23,FurelosBlancoLJBR23}. Our method falls into the latter category. We hypothesize that the appearance of erroneous labels in the noisy setting makes RMs challenging to learn for methods in the former category.

The learning of RMs from noisy labels has only been previously considered by \inlinecite{verginis2024joint}. Their work focuses on optimizing the noisy labelling function to model the perfect labelling function after thresholding, enabling the use of existing algorithms for RM learning. In contrast, our approach directly integrates noise into the RM learning procedure. Besides, \thesystem is orthogonal to the choice of the sensor model, so we could integrate a similar mechanism to achieve better results; however, their mechanism is less general since it assumes that in state $\state$, the perfect labelling function $\labelingfunction(\state)$ always returns the same label $\observationlabel$, making it unusable if the environment configuration (e.g., layout) changes between episodes.




Considering RMs with uncertainty more broadly, \inlinecite{CorazzaGN22} learn RMs that output stochastic rewards. 
\inlinecite{DBLP:conf/aips/DohmenTAB0V22} propose an algorithm for learning probabilistic RMs, which are stochastic in both the state-transition and the reward-transition functions. The exploitation of RMs with a noisy labelling function was previously studied by \citeauthor{LiCVKTM22} (\citeyear{LiCVKTM22,li2024reward}); indeed, we employ their \emph{independent belief updating} approach in the RM exploitation and discuss the \emph{thresholding} approach in the evaluation.

\section{Conclusions and Future Work}
\label{sec:conclusions}
In this paper, we have introduced $\thesystem$, a method for learning and exploiting RMs from noisy traces perceived by a RL agent.
We have shown that the method performs comparably to an algorithm where the RM is given in advance.
Being able to deal with noisy labelling functions enables more realistic applications of RM-learning. 

In future work, we plan to assess $\thesystem$'s performance in continuous domains such as \textsc{WaterWorld}~\cite{Karpathy15} and with autonomous embodied agents. Previous studies have demonstrated that incorporating an adversary into the RL training process enhances policy robustness \cite{pinto2017robust}; likewise, we aim to investigate whether RM learning from noisy examples is resilient against attacks on RMs \cite{nodari2023adversarial}. Finally, we strive to improve the system's scalability to learn RM with more than seven states. Learning hierarchies of RMs~\cite{lauffer2022learning,FurelosBlancoLJBR23} is a promising possibility since they enable learning smaller yet equivalent RMs.

\section*{Acknowledgements}
Research was sponsored by the Army Research Laboratory and was accomplished under Cooperative Agreement Number W911NF-222-0243. The views and conclusions contained in this document are those of the authors and should not be interpreted as representing the official policies, either expressed or implied, of the Army Research Laboratory or the U.S. Government. The U.S. Government is authorised to reproduce and distribute reprints for Government purposes notwithstanding any copyright notation herein.

This work was also supported by the UK EPSRC project EP/X040518/1. Roko Parać is supported by the UKRI Centre for Doctoral Training in Safe and Trusted AI (EPSRC Project EP/S023356/1). We thank Julian de Gortari Briseno for his help with the noisy sensor model.

\bibliographystyle{kr}
\bibliography{kr-sample}

\end{document}